\documentclass[letterpaper]{article}
\usepackage{aaai2026}
\usepackage{times,helvet,courier}
\usepackage[hyphens]{url}
\usepackage{natbib}
\usepackage{graphicx}
\usepackage{booktabs,multirow,tabularx}
\usepackage{amsmath,amssymb}
\usepackage{algorithm}
\usepackage{algpseudocode}
\usepackage{tikz}
\usetikzlibrary{arrows.meta,positioning,shapes.geometric,fit,backgrounds}
\nocopyright

\newcommand{\method}{\textsc{CoDA}}

\title{Learning from Consensus and Disagreement: Unsupervised On-Policy Self-Distillation with Minority-Trajectory Contrast}
\author{
    Jiaxin Guo\textsuperscript{\rm 1,\rm 2},
    Yanwei Yue\textsuperscript{\rm 1,\rm 2},
    Xuanbo Fan\textsuperscript{\rm 1,\rm 2},
    Chunyu Yang\textsuperscript{\rm 3},
    Yan Zhang\textsuperscript{\rm 1,\rm 2}\protect\thanks{Corresponding author}
}
\affiliations {
    \textsuperscript{\rm 1}School of Intelligence Science and Technology, Peking University\\
    \textsuperscript{\rm 2}State Key Laboratory of General Artificial Intelligence, Peking University, Beijing, China\\
    \textsuperscript{\rm 3}Ucap Cloud\\
}

\begin{document}
\maketitle

\begin{abstract}
On-policy self-distillation improves language-model reasoning by querying a teacher on states actually visited by the student. Recent methods create a powerful information asymmetry by exposing the teacher to privileged context, yet they fundamentally rely on external supervision---such as gold solutions or verifiers---to construct this advantage. We introduce \method{} (Consensus and Disagreement Alignment), a \textbf{fully unsupervised} framework that creates reliable privileged information entirely from the latent uncertainty structure of a model's own unlabeled rollouts. \method{} extracts two complementary signals. In the \textbf{positive branch}, answer-level consensus identifies a stable reasoning mode, which conditions a frozen self-teacher to provide dense distributional guidance on fresh student trajectories. However, because agreement does not guarantee correctness, positive-only distillation risks amplifying correlated errors into a false consensus. To break this harmful feedback loop, \method{} incorporates a \textbf{negative branch} that exploits disagreement: \textbf{minority trajectories} are treated as unstable alternatives and gently penalized via a reference-anchored, KTO-style calibration objective. This unpaired binary feedback provides robust regularization without requiring the strong assumption that the consensus is the absolute ground truth. Empirical evaluations on competition-level mathematical benchmarks demonstrate that \method{} \textbf{significantly improves reasoning}, outperforming self-generated baselines and effectively stabilizing training against erroneous consensus.
\end{abstract}

\begin{figure}[t]
    \centering
    \includegraphics[width=\columnwidth]{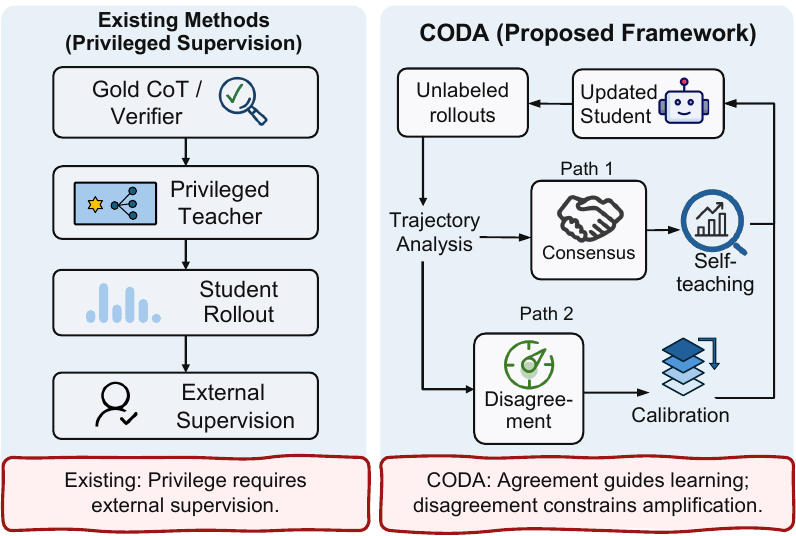}
    \caption{Contrast of two knowledge-amplification paradigms: existing privilege-based learning relies on externally supplied supervision, whereas \method{} constructs privileged context from agreement and uses disagreement calibration for robust unsupervised self-distillation.}
    \label{fig:intro}
    \end{figure}

\section{Introduction}

Long chains of thought have substantially improved language-model reasoning, but they also sharpen a fundamental train--test mismatch\cite{wei2022chain, agarwal2024policy}. A model trained on fixed expert traces learns under clean prefixes, yet at inference time must continue from its own imperfect decisions; a single early deviation can move it into states that offline supervision never covers. On-policy distillation mitigates this exposure gap by querying a teacher on trajectories sampled from the current student \citep{ross2011dagger,agarwal2024policy,minillm2024,distillm2024}. Recent on-policy self-distillation further shows that the same model can teach itself when the teacher receives \emph{privileged information} hidden from the student \citep{zhao2026opsd,penaloza2026pid,gates2026}. The resulting capability gap is not created by a larger teacher, but by an information asymmetry: conditioned on a solution or other task-specific evidence, a copy of the model can provide more informative token-level targets along the states actually visited by the student.

This perspective, however, exposes a critical supervision bottleneck. Existing privileged-information methods typically obtain their asymmetry from a reference chain of thought, a gold final answer, or a verifier-approved trajectory. Their apparent self-distillation therefore still fundamentally depends on an externally supplied object that tells the teacher what to know. Removing labels does more than remove a training target: it removes the source of the teacher's advantage. We ask a more basic question than how to distill without labels: \emph{can a reasoning model create reliable privileged information entirely from the uncertainty structure of its own unlabeled rollouts?}

Self-consistency offers a starting point for endogenous asymmetry. When independently sampled reasoning paths converge, their agreement reveals a stable policy mode \citep{wang2023self}. Exposing a representative consensus trajectory exclusively to a self-teacher creates privileged context, enabling dense distributional guidance on fresh student rollouts. However, agreement is not correctness. Correlated model errors can produce a \emph{false consensus}, and conditioning the teacher on this erroneous context rationalizes and propagates its bias\cite{xu2026tip,zheng2026scope}. Positive-only self-distillation thus closes a harmful feedback loop: it promotes mistakes to privileged information, converting sampling error into flawed-prefix self-amplification \citep{fu2026failure,guo2026mitigating, press2023measuring}.

Figure~\ref{fig:intro} contrasts existing paradigms with our approach. While prior methods rely on external supervision, we replace it with agreement-derived context and introduce disagreement calibration to constrain the amplification of unstable modes. Crucially, the same rollout set exposes not only agreement but also \emph{disagreement}. While consensus identifies stable reasoning modes, minority trajectories expose isolated, unstable alternatives. Exploiting this negative evidence does not require assuming the consensus is the absolute ground truth---a necessary concession when the model exhibits systematic biases. Instead, we impose a weaker, policy-relative constraint: an isolated alternative should not become more likely than it was under the initial policy. This parallels the ``lucky hit'' phenomenon in test-time reinforcement learning, where negative outcomes remain informative without absolute correctness guarantees \citep{ttrl2025}.

Based on this insight, \textbf{we introduce \method{} (\emph{Consensus and Disagreement Alignment}), a fully unsupervised framework with two complementary branches.} Given an unlabeled question, \method{} infers consensus by sampling diverse reasoning paths and grouping them by terminal answer. In the positive branch, a representative consensus trajectory conditions a privileged self-teacher, whose token distributions guide a fresh on-policy student rollout. In the negative branch, minority trajectories receive a reference-anchored, KTO-style undesirable-response objective relative to the frozen initial policy \citep{ethayarajh2024kto}. KTO is ideal here because its unpaired binary feedback gently discourages minority trajectories without constructing preference pairs or certifying the consensus as a correct winner. By bounding this penalization, the disagreement branch acts as a robust regularizer rather than a claim of semantic falsity, mitigating the risk of penalizing true answers under a biased policy. \textbf{Together, consensus creates an endogenous teacher--student information asymmetry, while disagreement prevents unstable self-generated modes from gaining probability unchecked.}

Our contributions are threefold:
\begin{itemize}
    \item We formulate \emph{fully unsupervised privileged on-policy self-distillation}: a model constructs self-teaching information asymmetry solely from the agreement structure of its unlabeled rollouts, requiring no gold solutions, reward models, or verifiers.
    \item We identify \emph{false-consensus amplification} as a failure mode of positive-only self-distillation and propose dual-signal alignment, combining consensus-conditioned distribution matching with reference-anchored calibration of disagreeing trajectories to suppress this bias.
    \item We empirically demonstrate that \method{} significantly improves reasoning, outperforming self-generated baselines. Extensive analysis validates our design effectively stabilizes training against noisy or erroneous consensus.
\end{itemize}

\section{Related Work}

\paragraph{On-policy distillation and privileged self-distillation.}
Traditional knowledge distillation suffers from exposure bias in autoregressive generation \citep{hinton2015distilling,ross2011dagger}. On-policy distillation mitigates this by aligning teacher and student distributions on student-generated sequences using various divergence metrics \citep{agarwal2024policy,minillm2024,distillm2024} or adaptive objectives to handle noisy feedback \citep{aopd2026,xu2026tip,zheng2026scope,paced2026}. Recently, privileged self-distillation removes the external teacher entirely by conditioning a frozen self-teacher on privileged context, such as gold solutions or externally validated consensus \citep{zhao2026opsd,penaloza2026pid,ye2026opcd,gates2026,unisd2026}. However, these methods fundamentally assume the privileged context is externally provided. We instead construct this asymmetry from unlabeled rollouts and address the robustness challenges of self-generated, noisy context.

\paragraph{Self-improvement from model-generated feedback.}
Models can self-improve using signals derived from their own generations, such as inference-time self-consistency \citep{wang2023self}, reward-guided self-training \citep{ttrl2025,sdzero2026,selfdistilledrlvr2026}, or preference optimization via paired \citep{rafailov2023direct,guo2026failures} and unpaired \citep{ethayarajh2024kto} objectives. In on-policy distillation, unreliable student prefixes are typically addressed via filtering or credit assignment \citep{fu2026failure,xu2026tip,zheng2026scope,shen2026antisd,shen2026credit}. Most closely related to our work is Multi-rollout OPD \citep{yu2026multirollout}, which exploits cross-rollout relationships to form peer-conditioned teacher signals. However, it relies on ground-truth correctness to partition successful and failed trajectories. Our setting is orthogonal: we operate entirely without ground-truth labels or verifiers, inferring supervision---and regularizing errors---solely from the uncertainty structure of unlabeled generations.

\section{Method}
\subsection{Problem Setup}
Let $\mathcal{D}_x=\{x_i\}_{i=1}^N$ be a collection of questions without solutions, final answers, preference labels, or executable rewards. We seek to improve an autoregressive reasoning policy $\pi_\theta$ using only text sampled from the policy itself. A completion $y=(y_1,\ldots,y_T)$ contains an unconstrained reasoning trace followed by a final answer. Its sequence probability factorizes as
\begin{equation}
 P_\theta(y\mid x)=\prod_{t=1}^{T}\pi_\theta(y_t\mid x,y_{<t}).
\end{equation}
At every update, the current policy generates two kinds of on-policy data. First, an \emph{evidence set} $\mathcal{Y}_\theta(x)=\{y^{(1)},\ldots,y^{(s)}\}$ is sampled to infer the latent agreement structure of the model's solutions. Second, after constructing privileged context from that set, the student independently samples a training trajectory $r\sim P_\theta(\cdot\mid x)$. Separating these roles is important: the evidence trajectories decide what information the teacher receives, whereas the fresh trajectory determines the prefixes on which teacher and student distributions are aligned.


The training signal must satisfy two constraints. It must be \emph{fully unsupervised}: no ground-truth answer may be used to accept, reject, or weight a training prompt. It must also be \emph{on-policy}: both the evidence set and the aligned trajectory are regenerated as $\pi_\theta$ evolves. The answer extractor $a(y)$ is used only to canonicalize the final answer for grouping; it does not judge correctness. Invalid or missing answers receive the special symbol $\bot$ and do not vote.

\begin{figure*}[t]
    \centering
    \includegraphics[width=\textwidth]{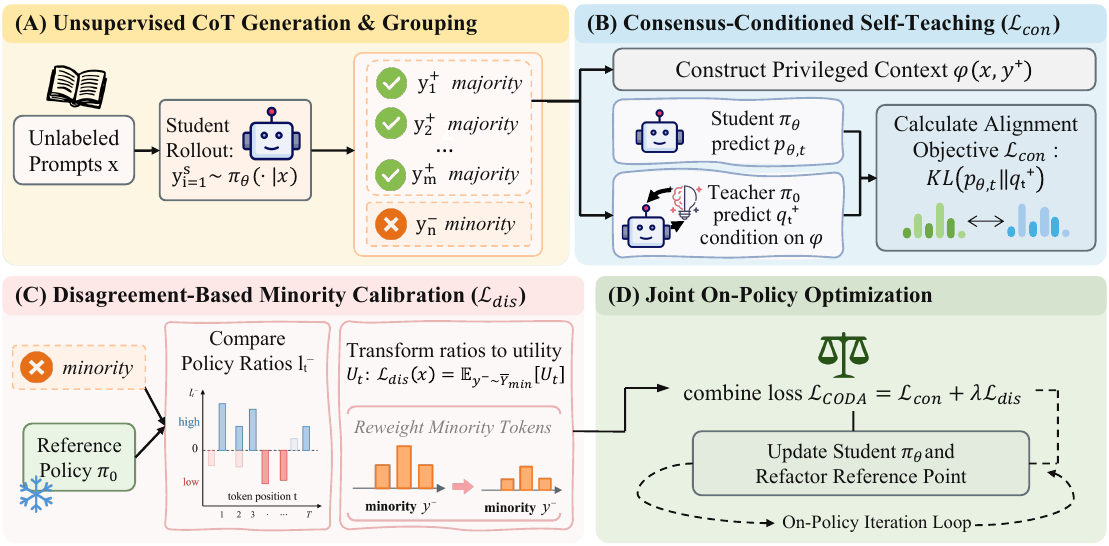}
    \caption{Overview of \method{}. (A) Unlabeled student rollouts are grouped by normalized final answer to identify consensus and disagreement trajectories. (B) A representative consensus trajectory provides privileged context for token-level teacher--student alignment on a fresh student rollout. (C) Minority trajectories are calibrated against the frozen reference policy, and (D) the two signals are combined in an on-policy update.}
    \label{fig:method}
    \end{figure*}

\subsection{From Self-Consistency to Privileged Context}
For every valid normalized answer $z$, we estimate its empirical support under the current policy,
\begin{equation}
 \widehat p_\theta(z\mid x)=\frac{1}{s}\sum_{j=1}^{s}
 \mathbb{I}\!\left[a(y^{(j)})=z\right].
\end{equation}
Let $c(z)=s\widehat p_\theta(z\mid x)$. We retain a prompt only when the largest answer group contains at least two trajectories, and define the possibly tied modal answer set
\begin{equation}
 \mathcal{Z}_{\mathrm{maj}}=\arg\max_z c(z), \qquad \max_z c(z)\geq 2.
\end{equation}
The modal and minority trajectory pools are then
\begin{equation}
 Y_{\mathrm{maj}}=\{y_i \mid a(y_i)\in\mathcal{Z}_{\mathrm{maj}}\},
\end{equation}
\begin{equation}
 Y_{\mathrm{min}}=\{y_i \mid a(y_i)\notin\mathcal{Z}_{\mathrm{maj}}\}.
\end{equation}
We select a representative $y^+ \in Y_{\mathrm{maj}}$ using a label-free selector $g$, without certifying its absolute correctness. To resolve ties in answer support, $g$ operates across the union of all tied modal pools. Since sequence length proxies the amount of reasoning exposed to the teacher, we empirically evaluate three selection rules for $g$: uniform random, shortest, and longest trajectory.

The resulting privileged context, $c^+=\phi(x,y^+)$, presents $y^+$ as a reference and prompts the teacher to solve the problem independently. This distinction prevents direct token imitation; while $y^+$ conditions the teacher's distribution, the alignment loss is strictly evaluated on the separately generated student trajectory $r$. We term this mechanism \emph{consensus-conditioned self-teaching}.


\subsection{Consensus-Conditioned Distribution Alignment}

The student samples an on-policy trajectory $r \sim P_\theta(\cdot \mid x)$ without observing $y^+$. At each prefix $r_{<t}$, the trainable student and the frozen privileged teacher yield next-token distributions under asymmetric information:
\begin{equation}
 p_{\theta,t}(v) = \pi_\theta(v \mid x, r_{<t}), \quad q^+_t(v) = \pi_0(v \mid \phi(x, y^+), r_{<t}).
\end{equation}
This formulation ensures the teacher's advantage stems solely from the consensus context while maintaining on-policy correction along student-visited states.

To align these distributions, we define a \emph{consensus-conditioned} objective using a generalized divergence $D_\alpha$:
\begin{equation}
 \mathcal{L}_{\mathrm{con}}(x) = \frac{1}{|M_r|} \sum_{t \in M_r} \min\!\left\{D_{\alpha}(q^+_t, p_{\theta,t}), \tau\right\},
\label{eq:consensus}
\end{equation}
where $M_r$ masks non-response tokens and $\tau$ clips excessive divergences to prevent formatting tokens from dominating the loss. The parameter $\alpha \in [0, 1]$ interpolates between forward KL ($\alpha=0$), generalized Jensen--Shannon ($0 < \alpha < 1$), and reverse KL ($\alpha=1$). In our experiments, we set $\alpha=0$ (forward KL) to transfer the teacher's full distribution.

While this positive branch provides dense guidance even from imperfect derivations, a false consensus can systematically bias $q^+_t$. This vulnerability necessitates a complementary mechanism whose validity does not rely on assuming $y^+$ is strictly correct.

\subsection{Reference-Anchored Disagreement Calibration}

For a minority trajectory $y^-\in Y_{\mathrm{min}}$, consensus alignment provides no explicit reason to suppress the behavior that produced it. Directly forming a preference pair $(y^+,y^-)$ would be overly strong, as modal membership does not guarantee $y^+$ is correct. We instead assign an unpaired undesirable label to $y^-$ using a reference-anchored, KTO-style objective \citep{ouyang2022training,ethayarajh2024kto}.

For a response token $y^-_t$, define the student--reference log-ratio:
\begin{equation}
 \ell_t^-(\theta)=\log\frac{\pi_\theta(y^-_t\mid x,y^-_{<t})}
 {\pi_0(y^-_t\mid x,y^-_{<t})}.
\end{equation}
Since our negative branch operates at the token level against a fixed reference policy, we fix the KTO reference point to zero and directly minimize the centered objective:
\begin{equation}
 \mathcal{L}_{\mathrm{dis}}(x)=
 \mathbb{E}_{\substack{y^-\sim\widetilde Y_{\mathrm{min}}\\t\sim M_{y^-}}}
 \left[\operatorname{softplus}\!\left(\beta\ell_t^-(\theta)\right)-\log 2\right],
\label{eq:disagreement}
\end{equation}
where $\widetilde Y_{\mathrm{min}}\subseteq Y_{\mathrm{min}}$ contains at most $K$ selected minority trajectories. 

By penalizing positive log-ratios, this objective reduces the relative likelihood of minority tokens. Crucially, the frozen $\pi_0$ anchor prevents indiscriminate likelihood minimization; it merely ensures that observed disagreement modes are not amplified beyond their initial probability.

Because minority membership is a weak signal---a minority trajectory might contain correct reasoning if the consensus is false--- we strictly treat this branch as a regularizer rather than a claim of semantic falsity, enforced via valid-answer filtering, length capping, and a small coefficient $\lambda$. 

Figure~\ref{fig:method} summarizes the resulting two-branch on-policy training loop, integrating consensus-conditioned self-teaching with reference-anchored disagreement calibration.

\subsection{Joint Objective and On-Policy Update}

The complete per-question objective combines the two complementary signals:
\begin{equation}
 \mathcal{L}_{\mathrm{CoDA}}(x)=
 \mathcal{L}_{\mathrm{con}}(x)+\lambda\mathcal{L}_{\mathrm{dis}}(x).
\label{eq:total}
\end{equation}
Here, $\mathcal{L}_{\mathrm{con}}$ extracts privileged guidance from repeated evidence, while $\mathcal{L}_{\mathrm{dis}}$ mitigates the risk of false consensus by regularizing isolated modes. 

Because all rollouts are continuously sampled from the active student $\pi_\theta$, both the constructed consensus context and the penalized minority modes dynamically evolve as the policy improves. To maintain signal quality, prompts lacking a repeated valid answer skip the update entirely. Crucially, this joint on-policy loop operates strictly without ground-truth labels, extracting both learning and regularization signals purely from the model's internal rollout structure.


\section{Experiments}
\subsection{Datasets and Metrics}
We train on mathematical reasoning problems from \textbf{OpenThoughts}~\cite{guha2025openthoughts}. While supervised baselines utilize the provided expert solutions, \method{} constructs its training signal entirely from the problem prompts and its own on-policy rollouts. We evaluate on five competition-level benchmarks: \textbf{AIME 2024, 2025, 2026}, \textbf{HMMT 2025}, and \textbf{AMO-Bench}. 
For evaluation, we report \textbf{avg@16} (sample-average accuracy across 16 generations).

\subsection{Baselines}
We compare \method{} against the unmodified \textbf{base model} and two groups of post-training baselines:
(1) \textbf{Supervised} methods using expert trajectories or gold labels: SFT (off-policy imitation), OPSD~\cite{zhao2026opsd} (on-policy self-distillation conditioned on gold solutions), and GRPO~\cite{shao2024deepseekmath} (optimizing group-relative advantages via binary correctness rewards).
(2) \textbf{Self-Generated} methods: \textbf{SFT-Self (oracle-filtered)\cite{wang2023self}} (SFT on self-generated trajectories filtered by ground truth), TTRL\cite{ttrl2025} (rewarding answer agreement without gold labels), \method{}--Consensus Only (our positive branch alone), and \method{}--Full. Notably, \method{}--Full operates entirely without external supervision or oracle filtering.

\begin{table*}
    \centering
    \small
    \resizebox{\textwidth}{!}{%
    \begin{tabular}{cllcccccc}
    \toprule
    Model & Setting & Method & AIME24 & AIME25 & AIME26 & AMO & HMMT25 & Avg. \\
    \midrule
    \multirow{8}{*}{Qwen3-4B}
    &  & Base Model & 74.38 & 65.83 & 65.21 & 12.12 & 42.29 & 51.97 \\
    & \multirow{3}{*}{Supervised} & SFT & 72.71 & 68.12 & 67.29 & 11.38 & 40.00 & 51.90 \\
    & & OPSD & \textbf{75.42} & 68.75 & \textbf{69.58} & \textbf{14.50} & \textbf{45.83} & \textbf{54.82} \\
    & & GRPO & 75.21 & \textbf{70.00} & 67.71 & 13.88 & 45.21 & 54.40 \\
    \cmidrule(lr){2-9}
    & \multirow{4}{*}{Self-Generated} & SFT-Self (oracle-filtered) & \textbf{75.00} & 66.46 & 66.88 & 13.62 & 45.21 & 53.43 \\
    & & TTRL & 74.38 & 66.25 & 66.04 & 13.00 & 43.96 & 52.73 \\
    & & \method{}--Consensus Only & 74.58 & 67.08 & 67.92 & 12.88 & \textbf{46.04} & 53.70 \\
    & & \method{}--Full & 74.58 & \textbf{68.96} & \textbf{70.21} & \textbf{14.50} & 44.58 & \textbf{54.57} \\
    \midrule
    \multirow{8}{*}{Qwen3-1.7B}
    &  & Base Model & 48.33 & 37.08 & 36.67 & 3.50 & 23.12 & 29.74 \\
    & \multirow{3}{*}{Supervised} & SFT & 51.25 & 37.92 & 40.83 & 4.38 & 26.25 & 32.13 \\
    & & OPSD & \textbf{57.29} & \textbf{42.29} & \textbf{46.88} & 4.88 & \textbf{27.71} & \textbf{35.81} \\
    & & GRPO & 51.10 & 39.58 & 40.83 & \textbf{5.50} & 26.46 & 32.69 \\
    \cmidrule(lr){2-9}
    & \multirow{4}{*}{Self-Generated} & SFT-Self (oracle-filtered) & 52.08 & 38.75 & 38.33 & 4.50 & 25.00 & 31.73 \\
    & & TTRL & 50.00 & 38.33 & 37.92 & \textbf{5.00} & 25.83 & 31.42 \\
    & & \method{}--Consensus Only & 53.33 & \textbf{42.08} & 44.38 & 3.75 & 28.33 & 34.37 \\
    & & \method{}--Full & \textbf{56.67} & 41.67 & \textbf{47.29} & 4.75 & \textbf{30.21} & \textbf{36.12} \\
    \bottomrule
    \end{tabular}
    }
    \caption{Main reasoning results on Qwen3-4B and Qwen3-1.7B (accuracy, \%). ``Avg.'' is the unweighted average over the five benchmarks. Bold indicates the best result within each setting (Supervised vs. Self-Generated). SFT-Self uses oracle correctness filtering and is included as a diagnostic self-imitation reference, not as a fully unsupervised method.}
    \label{tab:main}
\end{table*}

\begin{table*}[t]
\centering
\begin{tabular}{lcccccc}
\toprule
Selector & AIME24 & AIME25 & AIME26 & AMO & HMMT25 & Avg. \\
\midrule
Random & 55.00 & 41.46 & 46.67 & 4.00 & 29.17 & 35.26 \\
Shortest & \textbf{56.67} & \textbf{41.67} & \textbf{47.29} & \textbf{4.75} & 30.21 & \textbf{36.12} \\
Longest & 54.79 & 41.04 & 46.25 & 3.88 & \textbf{30.42} & 35.28 \\
\bottomrule
\end{tabular}
\caption{Comparison of trajectory selectors from the modal answer pool on Qwen3-1.7B (accuracy, \%). Selecting the shortest trajectory consistently yields the best overall performance (best values in bold).}
\label{tab:selector}
\end{table*}

\subsection{Implementation Details}
Experiments use instruction-tuned \textbf{Qwen3-1.7B} and \textbf{Qwen3-4B}~\cite{qwen3report}. We apply LoRA~\cite{hu2022lora} ($r=64$, $\alpha=128$) to all linear projection modules. Training runs for one epoch using bfloat16, FlashAttention-2, and vLLM on 8 NVIDIA H100 GPUs, with a $5\times10^{-6}$ learning rate and 0.1 gradient clipping.

For \method{}, each prompt generates $s=10$ candidate trajectories (temperature 1.3, top-$p=0.95$). Prompts require at least two valid responses sharing a modal answer to proceed. To construct the privileged context, Qwen3-1.7B selects the shortest modal trajectory (analyzed in Section~\ref{sec:selector-analysis}), while Qwen3-4B samples one randomly. 

The adapter-disabled initial policy serves as both the frozen teacher and reference anchor. The consensus branch aligns forward KL on a fresh student rollout (max 2,048 tokens), clipping token-level divergences at 0.05. The disagreement calibration branch ($\beta=0.1$) limits training to at most one minority trajectory (max 1,024 tokens) per prompt.

\section{Results}
\subsection{Overall Performance}
Table~\ref{tab:main} evaluates post-training methods on Qwen3-4B and Qwen3-1.7B. We compare fully unsupervised approaches (\method{}, TTRL) against supervised references (SFT, OPSD, GRPO) and an oracle-filtered diagnostic (SFT-Self) to determine the efficacy of unlabeled rollout structures.

\paragraph{\method{} improves reasoning without correctness supervision.}
\method{}--Full achieves the strongest average performance among fully unsupervised methods across both model scales. On Qwen3-4B, it improves the base model by 2.60 points and outperforms TTRL by 1.84 points, reaching 54.57\%---only 0.25 points behind supervised OPSD. On Qwen3-1.7B, the trend amplifies: \method{} improves the base model by 6.38 points, exceeding supervised OPSD by 0.31 points. These broad improvements across multiple benchmarks confirm that repeated answer agreement supplies highly effective privileged context, even without reference solutions or ground-truth labels.

\paragraph{Minority calibration complements consensus alignment.}
Comparing \method{}--Consensus Only with \method{}--Full isolates the contribution of disagreement calibration. At 1.7B, the minority-trajectory objective increases average accuracy from 34.37\% to 36.12\%, improving four out of five benchmarks. The 4B model shows a consistent average gain (53.70\% to 54.57\%), particularly on more challenging evaluations like AIME 2025, AIME 2026, and AMO. While heuristic negative supervision naturally introduces minor trade-offs (e.g., on HMMT 2025), the aggregate gains across both scales demonstrate that penalizing isolated modes effectively counteracts the self-amplification risks of positive-only distillation.

\paragraph{Relation to oracle-filtered self-imitation and supervised references.}
\method{}--Full significantly surpasses the oracle-filtered SFT-Self baseline. This gap indicates that our improvements stem from dense, consensus-conditioned teacher guidance and disagreement regularization, rather than merely filtering plausible solutions. By approaching (at 4B) or exceeding (at 1.7B) the aggregate accuracy of supervised methods, the results demonstrate that jointly exploiting endogenous agreement and disagreement is a highly competitive alternative to correctness-filtered self-training.

\subsection{Selecting Privileged Context from Consensus}
\label{sec:selector-analysis}
Answer-level self-consistency identifies a modal pool, but the positive branch still requires a representative trajectory to instantiate the teacher's privileged context. We therefore ask whether this choice should be random or should favor a particular response length.

\paragraph{Final performance.}
Answer-level agreement identifies a set of candidate trajectories, but it does not specify which member should be revealed to the teacher. We evaluate uniform random selection against two deterministic, label-free alternatives: the shortest and longest trajectories in $Y_{\mathrm{maj}}$. Table~\ref{tab:selector} reports final sample-average accuracy for Qwen3-1.7B. Selecting the shortest modal trajectory produces the highest average accuracy, improving over random selection by 0.86 points and over the longest selector by 0.84 points. The improvement appears on four of five benchmarks, including all three AIME evaluations and AMO. Thus, the gain from consensus-conditioned self-teaching does not require exposing the teacher to the longest available derivation; a concise representative is at least as informative and is less likely to introduce superfluous reasoning into the privileged context.

\begin{figure}[t]
    \centering
    \includegraphics[width=1\columnwidth]{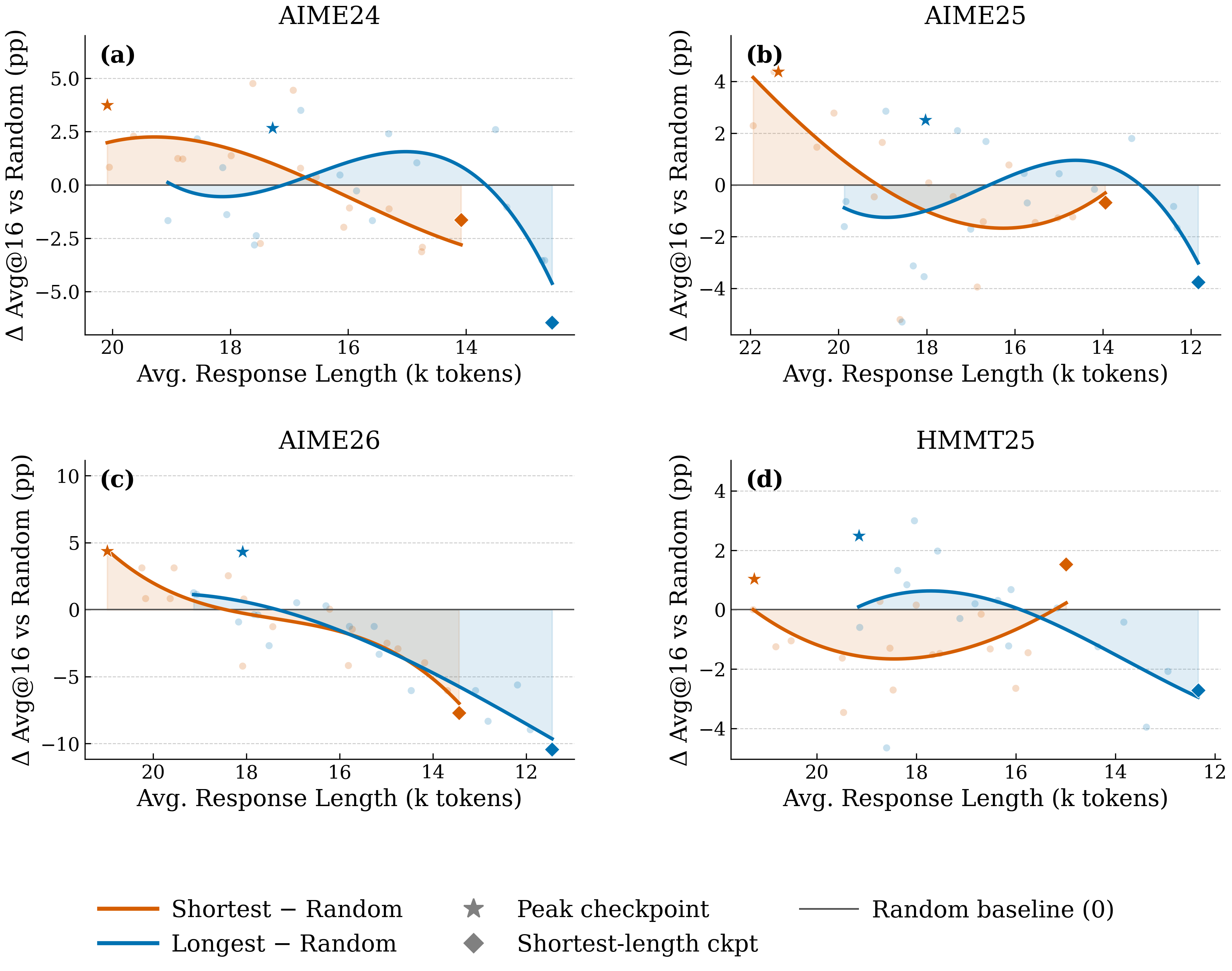}
    \caption{Accuracy difference of shortest vs. longest modal-trajectory selection relative to a random baseline. Notably, the shortest selector yields robust positive gains when the overall average response length is long (left side of panels), but this advantage diminishes as responses become shorter.}
    \label{fig:selector}
\end{figure}

\paragraph{Response length and accuracy.}
Figure~\ref{fig:selector} further traces the performance difference relative to uniform random selection as the average response length changes over training. The horizontal zero line represents the random-selector baseline. Notably, the shortest selector yields the most substantial positive gains when the overall average response length is long. However, as the model's responses naturally become shorter over training, this advantage diminishes and occasionally turns negative.

This dynamic reveals a clear structural trade-off. When the policy generates verbose trajectories, selecting the most concise consensus effectively strips away redundant exploration and correlated errors. Conversely, once outputs are already concise, strictly enforcing the shortest selection may inadvertently discard necessary reasoning steps. This observation rules out a simple ``more tokens is better'' hypothesis: the ideal privileged context must capture the stable solution mode while minimizing noisy verbosity.

\begin{figure}[!ht]
\centering
\includegraphics[width=\columnwidth]{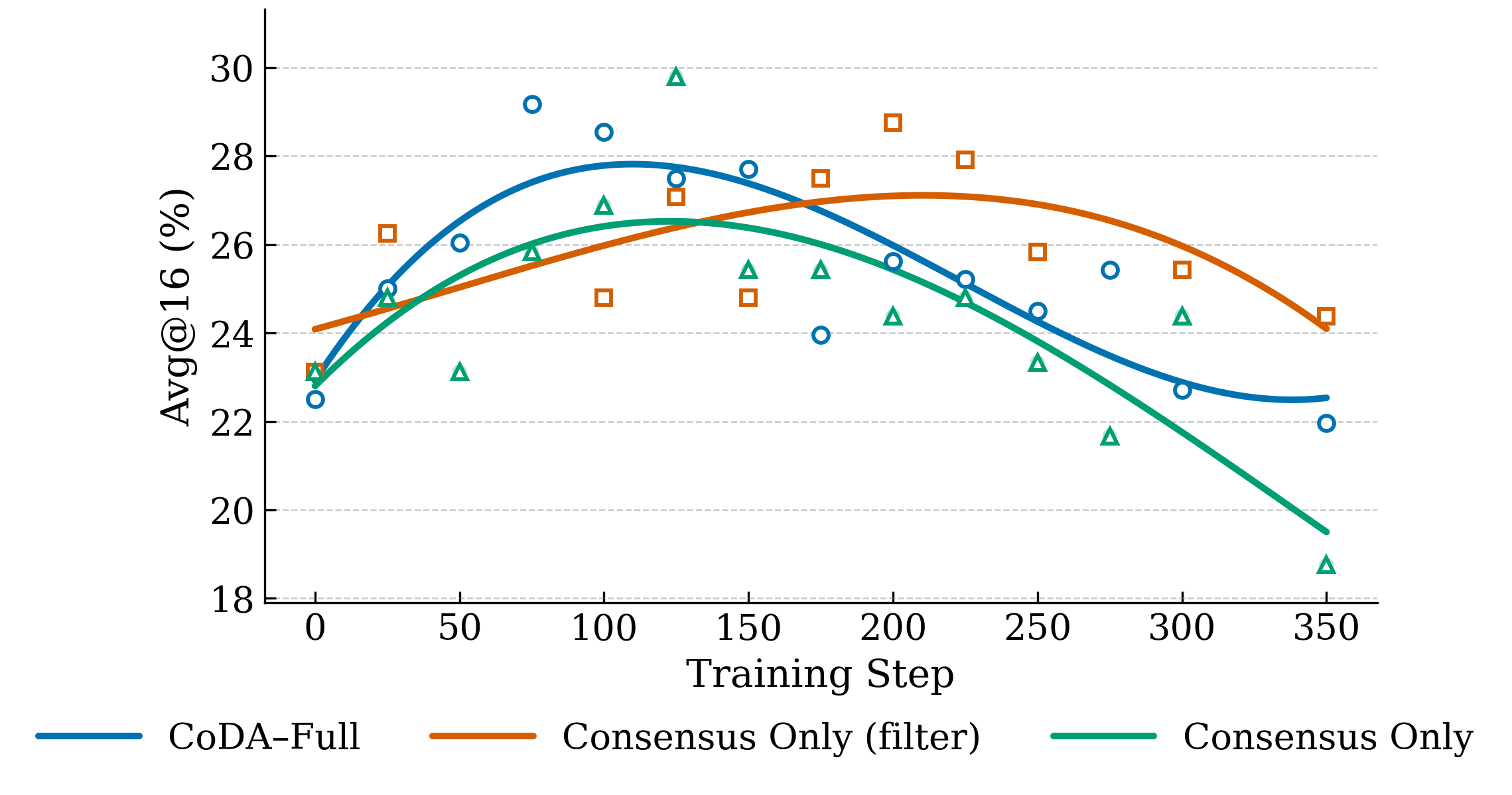}
\caption{The sample-average accuracy over training steps. Consensus Only improves early but undergoes the largest late-stage decline. Oracle filtering does not yield the highest peak, yet produces the most stable trajectory. \method{}--Full displays an intermediate pattern: disagreement calibration reduces the collapse.}
\label{fig:curve}
\end{figure}

\subsection{False Consensus and Training Stability}
\label{sec:false-consensus}
Consensus provides accessible privilege, but modal membership does not guarantee correctness. We investigate whether erroneous consensus accumulates during training and whether disagreement calibration mitigates this failure mode by comparing \method{}--Full against consensus-only alignment, both with and without an oracle quality filter (used strictly as a diagnostic using gold answers).

\begin{table*}[t]
\centering
\begin{tabular}{llcccccc}
\toprule
Model & Variant & AIME24 & AIME25 & AIME26 & AMO & HMMT25 & Avg. \\
\midrule
\multirow{3}{*}{Qwen3-4B}
& \method{}--Full & 74.58 & \textbf{68.96} & \textbf{70.21} & \textbf{14.50} & 44.58 & \textbf{54.57} \\
& Consensus Only & 74.58 & 67.08 & 67.92 & 12.88 & \textbf{46.04} & 53.70 \\
& Consensus Only + Oracle Filter & \textbf{74.79} & 66.25 & 69.58 & 13.25 & 45.62 & 53.90 \\
\midrule
\multirow{3}{*}{Qwen3-1.7B}
& \method{}--Full & \textbf{56.67} & \textbf{41.67} & 47.29 & \textbf{4.75} & 30.21 & \textbf{36.12} \\
& Consensus Only & 54.79 & 41.04 & 46.25 & 3.88 & \textbf{30.42} & \textbf{35.28} \\
& Consensus Only + Oracle Filter & 54.37 & 40.62 & \textbf{48.12} & 4.00 & 28.75 & 35.17 \\
\bottomrule
\end{tabular}
\caption{False-consensus analysis (sample-average accuracy, \%). ``Consensus Only'' uses $\mathcal{L}_{\mathrm{con}}$ without disagreement calibration. ``+ Oracle Filter'' additionally discards prompts whose modal answer is incorrect, using gold answers only for this diagnostic. Bold indicates the best value within each model block.}
\label{tab:false-consensus}
\end{table*}
\paragraph{Final performance.}
To isolate the effect of modal uncertainty, Table~\ref{tab:false-consensus} compares \method{}--Full, Consensus Only, and its oracle-filtered variant (which removes prompts with incorrect modal answers). Interestingly, perfect filtering does not consistently improve final average accuracy, shifting it from 53.70\% to 53.90\% at 4B and 35.28\% to 35.17\% at 1.7B. Thus, oracle filtering alone is insufficient to boost peak performance; its primary value is revealing how noisy consensus impacts optimization dynamics.

\paragraph{Training dynamics.}
Figure~\ref{fig:curve} tracks HMMT25 accuracy over training steps. While all variants initially improve, unfiltered Consensus Only suffers a severe late-stage decline. Conversely, oracle filtering maintains a smooth trajectory. This comparison confirms the false-consensus hazard: repeatedly conditioning the teacher on incorrect modal trajectories propagates misleading distributional targets, whereas removing them prevents error feedback.

\paragraph{Effect of disagreement calibration.}
\method{}--Full exhibits an intermediate behavior: its performance declines more gradually than unfiltered Consensus Only rather than collapsing. This matches our expectation that the KTO-style disagreement loss constrains the amplification of unstable modes despite imperfect modal contexts. While it does not fully match the stability of the gold-standard oracle filter, disagreement calibration successfully recovers a major share of this robustness without requiring any correctness labels.

\subsection{Effect of Consensus Sample Size}
\label{sec:sample-size}
The consensus estimator depends on the number of trajectories sampled per prompt. A larger evidence set can reduce sampling variance and make the modal answer more representative of the policy distribution, but its generation cost grows linearly with the sample count. We study this trade-off on Qwen3-1.7B using random selection from the modal pool and vary $s\in\{5,10,20\}$ while keeping the remaining training configuration fixed.

\begin{table}[t]
\centering
\begin{tabular}{lc}
\toprule
Consensus samples $s$ & Avg. (\%) \\
\midrule
5 & 35.05 \\
10 & 35.26 \\
20 & 35.41 \\
\bottomrule
\end{tabular}
\caption{Effect of the consensus sample using random modal-trajectory selection. The $s=20$ final evaluation is still in progress and is not estimated from training metrics.}
\label{tab:sample-size}
\end{table}

\begin{figure}[t]
\centering
\includegraphics[width=\columnwidth]{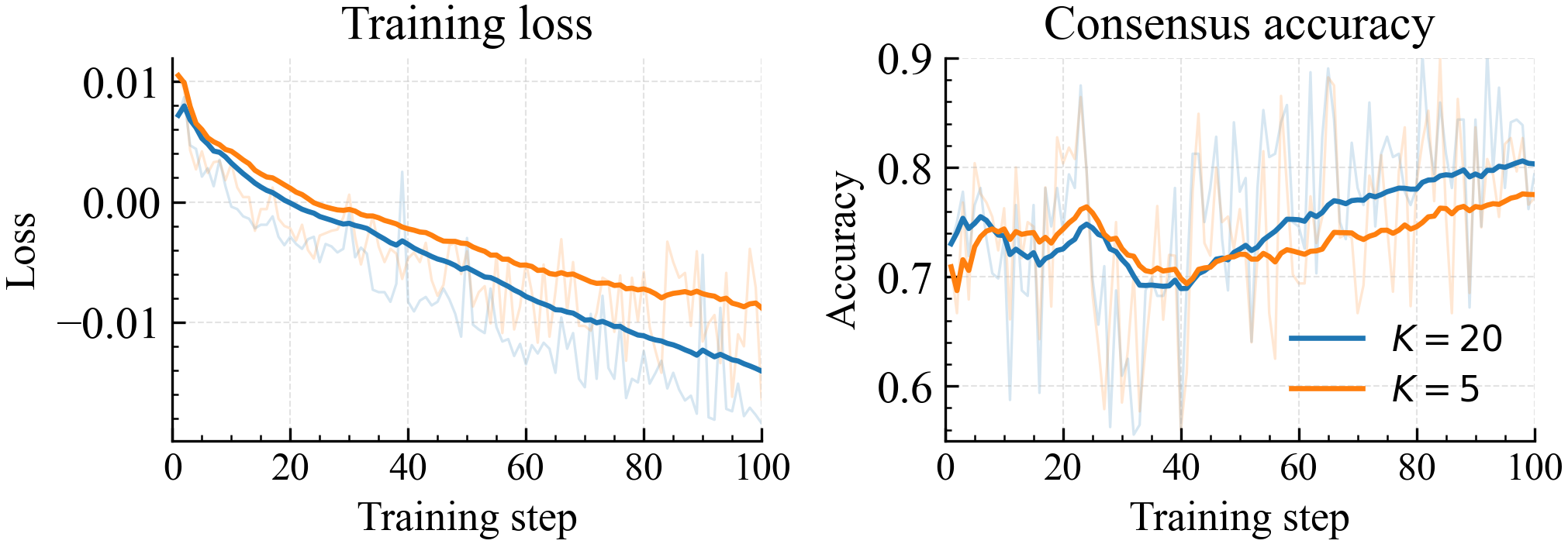}
\caption{Effect of consensus sample count ($s$) on Qwen3-1.7B training dynamics (legend $K$ denotes $s$). A larger evidence set ($s=20$) yields lower training loss and higher consensus accuracy, reflecting a less noisy privileged signal. }
\label{fig:sample-size}
\end{figure}

\paragraph{Optimization behavior.}
Figure~\ref{fig:sample-size} compares the two endpoints, $s=5$ and $s=20$. As training proceeds, the larger evidence set increasingly separates from $s=5$: it reaches a lower consensus-alignment loss and a higher consensus accuracy. The widening gap suggests that additional rollouts reduce finite-sample noise in the modal-answer estimate. Consequently, the privileged teacher is conditioned on a more stable consensus signal, producing more coherent distributional targets and allowing the student to fit the training prompts more rapidly. This result supports increasing $s$ when optimization stability is the primary concern.

\paragraph{Diminishing returns and capability limits.}
The available final evaluations nevertheless show only a modest improvement from $s=5$ to $s=10$ (35.05\% to 35.26\%). This small gain is consistent with \emph{consensus saturation}. A moderate sample count already resolves many prompts for which the correct answer is the dominant policy mode; additional samples mainly affect borderline cases or suppress a small number of sampling outliers. More samples cannot correct a systematic error when the model repeatedly follows the same incorrect reasoning pattern, because the wrong answer remains the modal outcome even under a more accurate estimate. Thus, increasing $s$ improves estimation of the model's current answer distribution, but does not remove the capability bound of that distribution.

\paragraph{Accuracy--efficiency trade-off.}
Increasing $s$ from 5 to 20 requires four times as many generated reasoning trajectories per prompt. The curves indicate a cleaner training signal, but the completed final results do not yet establish a commensurate generalization gain. We therefore use $s=10$ as the default: it provides a stronger consensus estimate than $s=5$ while avoiding the full generation cost of $s=20$. The takeaway is that larger evidence sets accelerate and stabilize consensus-conditioned optimization, whereas moderate $s$ offers a more favorable accuracy--compute trade-off.

\section{Conclusion}
We introduced \method{}, a fully unsupervised on-policy self-distillation framework that enhances language model reasoning without gold labels or oracle filtering. Rather than blindly trusting the most frequent generation, \method{} constructs privileged context from answer-level consensus while regularizing false consensus by penalizing minority trajectories via a reference-anchored objective. Evaluations on competition-level mathematical benchmarks demonstrate that this dual-signal alignment significantly outperforms positive-only baselines, prevents late-stage training collapse, and rivals fully supervised methods. Ultimately, \method{} establishes that the internal uncertainty structure of unlabeled rollouts provides a robust foundation for self-improvement.

\bibliography{references}
\end{document}